\documentclass[10pt,twocolumn,letterpaper]{article}

\usepackage{iccv}
\usepackage{times}
\usepackage{epsfig}
\usepackage{graphicx}
\usepackage{amsmath}
\usepackage{amssymb}
\usepackage{graphicx}
\usepackage{amsmath}
\usepackage{amssymb}
\usepackage{booktabs}
\usepackage{adjustbox}
\usepackage{url}

\usepackage[pagebackref=true,breaklinks=true,letterpaper=true,colorlinks,bookmarks=false]{hyperref}
\renewcommand*\backref[1]{\ifx#1\relax \else (Cited on page #1) \fi}
\newcommand{\squeezeup}{\vspace{-2mm}}

\usepackage[capitalize]{cleveref}
\crefname{section}{Sec.}{Secs.}
\Crefname{section}{Section}{Sections}
\Crefname{table}{Table}{Tables}
\crefname{table}{Tab.}{Tabs.}

\iccvfinalcopy 

\def\iccvPaperID{7586} 

\begin{document}

\title{NeAT: Neural Artistic Tracing for Beautiful Style Transfer}

\author{Dan Ruta$^1$, Andrew Gilbert$^1$, John Collomosse$^{1, 2}$, Eli Shechtman$^{2}$, Nicholas Kolkin$^{2}$\\
$^1$University of Surrey, $^2$Adobe Research\\
}

\ificcvfinal\thispagestyle{empty}\fi


\twocolumn[{%
\renewcommand\twocolumn[1][]{#1}%
\maketitle
\vspace{-20pt}

    \hspace{-0.025\linewidth}\includegraphics[width=1.05\linewidth]{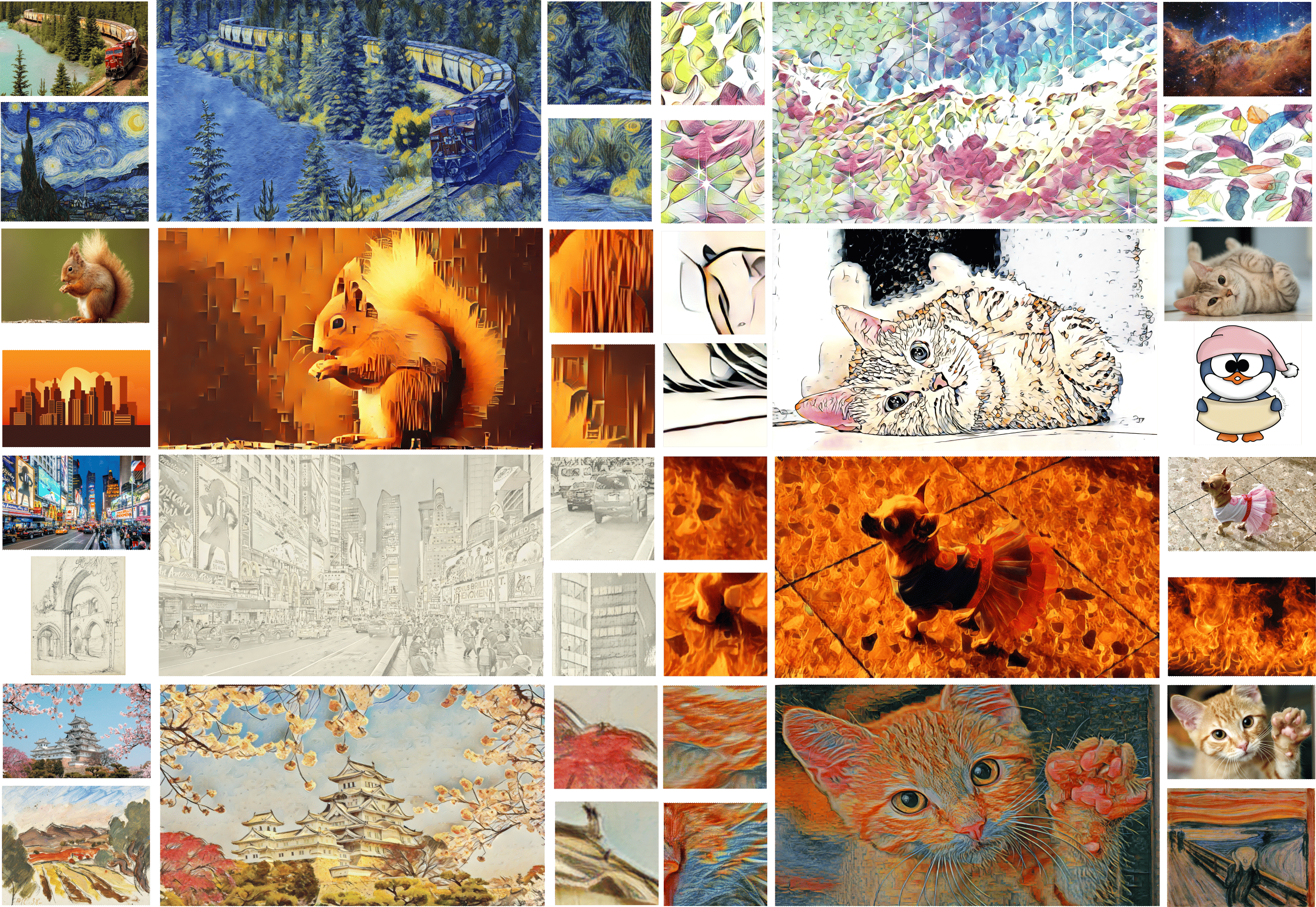}
\begin{center}
    \centering
Figure 1. Style transfer results using NeAT, trained on BBST-4M. Zoomed in areas are shown in the middle columns.
\end{center}%
\vspace{9pt}
}
]
\setcounter{figure}{1}

\begin{abstract}
\vspace{-0.35cm}
Style transfer is the task of reproducing the semantic contents of a source image in the artistic style of a second target image. In this paper, we present NeAT, a new state-of-the art feed-forward style transfer method. We re-formulate feed-forward style transfer as image editing, rather than image generation, resulting in a model which improves over the state-of-the-art in both preserving the source content and matching the target style. An important component of our model's success is identifying and fixing "style halos", a commonly occurring artefact across many style transfer techniques. In addition to training and testing on standard datasets, we introduce the BBST-4M dataset\footnote{\url{https://github.com/DanRuta/NeAT}}, a new, large scale, high resolution dataset of 4M images. As a component of curating this data, we present a novel model able to classify if an image is stylistic. We use BBST-4M to improve and measure the generalization of NeAT across a huge variety of styles. Not only does NeAT offer state-of-the-art quality and generalization, it is designed and trained for fast inference at high resolution. 
\vspace{-0.4cm}
\end{abstract}

\section{Introduction}
\label{sec:intro}

Since the introduction of techniques for  neural style transfer (NST) by Gatys \etal \cite{gatys}, there has been an explosion of research driving style transfer with deep neural networks. 
Initially performed via an optimization based approach, interest has gradually shifted to parametric and feed-forward approaches. 
The primary drawback of optimization approaches are their lengthy run-times, limiting their practical use. The research field has thus focused on advancing the quality and style diversity of feed-forward approaches to match or exceed that of optimization models, whilst maintaining practical run-times.  First, by adapting models to contain stylization capabilities for multiple styles \cite{Karayev2014,Collomosse2017}, and eventually arbitrary styles \cite{adain,wct}.

We set out to achieve 3 main objectives: \textbf{A)} state-of-the-art generality to both modern and classical styles; \textbf{B)} efficient scaling to high resolution images; and \textbf{C)} state-of-the-art visual stylization quality. We achieve all three of these objectives at the same time. Our technical contributions are:

\noindent \textbf{1.} - Reformulation of NST to modify pixels in a content image rather than to generate a brand new image
(contrasting with prior work). \textit{This achieves}: Better detail preservation from the content image, measured via our user
study, and visualized in Fig \ref{fig:deltasAblation}.

\noindent \textbf{2.} A new Sobel-based feature-complexity guided discriminator loss for a better matched stylization learning signal. \textit{This achieves}: Solving a long-standing issue - style halos present in prior NST. We are first to do so (Figs. \ref{fig:arch_sobel},\ref{fig:halos}).

\noindent \textbf{3.} Big Beautiful Style Transfer (BBST-4M) - a novel dataset of 4 million very high resolution images for the NST community. WikiArt and MSCoco ($\sim$80k images) typically used in NST, limit techniques to low resolution and focus on traditional
fine-art styles. \textit{This achieves}: High resolution output, and improved style diversity/quality measured in Sec. \ref{sec:generalizability} through a user
study comparing training on WikiArt+MSCoco vs. BBST-4M. Users prefer BBST-4M, especially for modern styles. We filter the style/content subsets of this dataset with the assistance of a novel dual-branch ViT \cite{vit} and ALADIN \cite{aladin} model trained to predict how artistic an image is. We release both this stylistic prediction model, and the BBST-4M dataset.

\section{Related Work}

\begin{figure*}
    \begin{center}
    \includegraphics[width=0.92\linewidth]{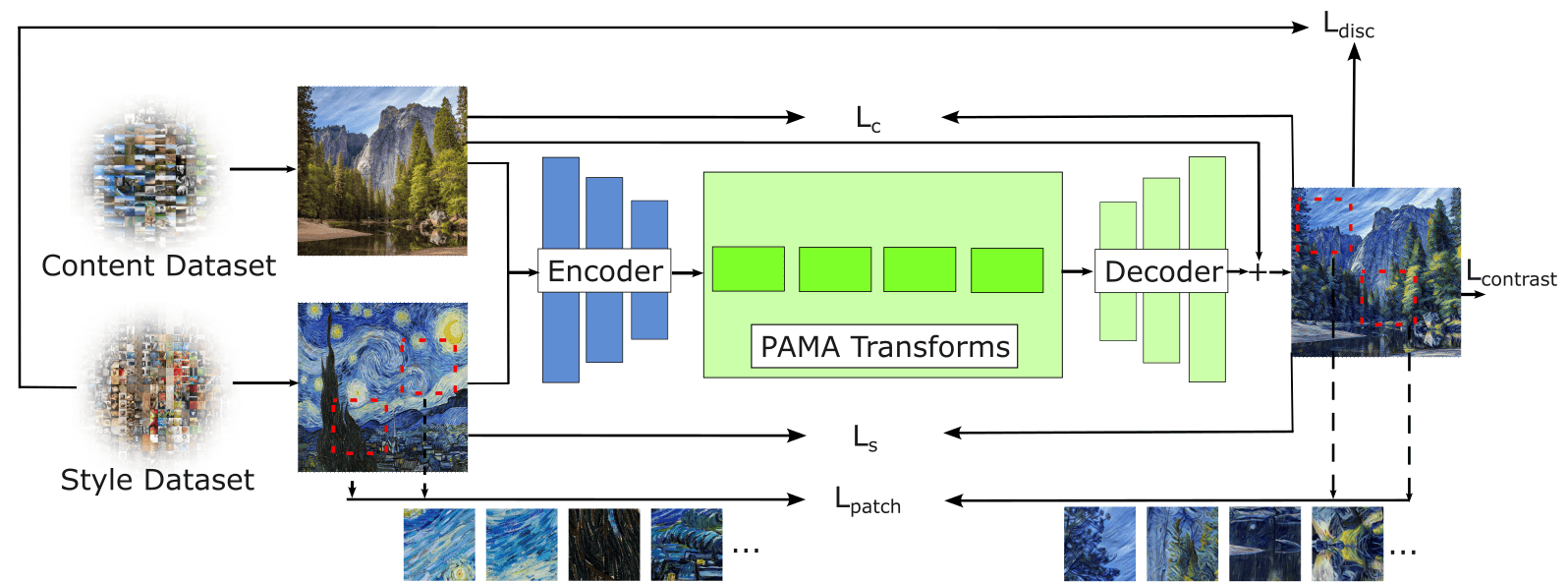}    
    \end{center}
   \squeezeup
   \squeezeup
   \caption{ Basic architecture diagram. The green modules represent the trainable parameters. For clarity, the discriminator modules and contrastive projection heads are not shown. The contrastive loss is computed 1) between the stylized output and the target style, and 2) between the stylized images only \cite{contraAST} - a group of contrastive losses anchoring on the same style regardless of content, and a group of contrastive losses anchoring on the same content regardless of style. 
   We also leverage several common identity losses between the stylized image and the respective style/content images from the datasets - computed as a Gatys loss \cite{gatys}.}
    \label{fig:arch}
\end{figure*}

Great stylization quality has been achieved in literature by focusing on the aligning the statistics of features extracted by frozen pretrained networks in the final output image and the target style image \cite{gatys,adain,strotts,nnst}. However, as with classical computer vision techniques, these approaches are limited by human design, and are likely sub-optimal, in quality and runtime.

More recent works have started exploring the idea of shifting this design process to the actual model via transformations learned through attention mechanisms. SANet \cite{sanet} introduced a style attentional module to perform style transfer in a feed-forward approach. 
Improving on local style information compared to Avatar-Net \cite{avatarnet}, they 

combine global and local style patterns extracted from different VGG layers, by normalizing the content/style feature maps, and computing attention between them via learnt 1x1 convolutions \cite{sagan}. 
PAMA \cite{pama} introduces iterative alignment of the content with the target style across multiple attention modules, improving quality and reducing inconsistencies of style across similar areas in the content image.

ContraAST \cite{contraAST} expands SANet by injecting information about the entire distribution of training data via domain-level adversarial losses. This is implemented by a discriminator acting on the stylized image, aiming to classify these outputs as \textit{fakes} and images from the style dataset as \textit{reals}. ContraAST also introduces contrastive losses between the stylized images. One encourages several stylized images with the same target style to stylistically match. Another encourages several stylizations of the same source content image but different styles to share the same content. These losses shift the burden of designing feature spaces and comparing them from human design into the learning process.

CAST \cite{cast} improves on the contrastive loss from ContraAST by including ground truth images in the contrastive pairs, expanding the loss to compare the generated images to images in the dataset.

This work also leverages several important ideas from the style transfer literature. In image translation literature, Swapping Auto-encoders \cite{sae} introduce a powerful discriminator based on patch co-occurrence. While it was originally used for photo-realistic style transfer, we find it equally powerful for artistic stylization.
Another important concept is increasing the effective batch size used in contrastive loss; as it has been shown that the value of contrastive loss scales well with increasing numbers of negative examples \cite{simclr}. However, in non-trivial models optimally high batch sizes are difficult to fit into VRAM. Logit accumulation, introduced in ALADIN \cite{aladin} enables compatibility of the gradient accumulation technique to models trained using contrastive loss. This works by iteratively accumulating the model's output logits without model gradients, performing the large batch-wise contrastive loss, then iteratively propagating the sample-wise gradients back through the model in sub-batches, this time using gradients from re-forwarding the respective sub-batch images through the model. Other similar accumulation approaches have been used in works such as Listwise loss \cite{listwise}, and ArtFlow \cite{artflow}.

Recently, diffusion based generative models \cite{stability_ldm,dalle2,makeascene} have attracted much attention in the research community. These models exhibit impressive quality in novel image generation. Through extensions such as textual inversion \cite{textualinversion} and Dreambooth \cite{dreambooth} existing models can be finetuned to generate specific concepts or styles. SDEdit\cite{sdedit} offers a mechanism to alter an existing image, although without much finegrained control over content preservation. Prompt2Prompt \cite{prompt2prompt} offers a mechanism to improve content preservation while editing, but has unpredictable failure cases and works best on generated images rather than real photos. Imagic \cite{imagic}, and Unitune \cite{unitune}, finetune the diffusion model for editing a specific image, making them extremely time and compute intensive. While these methods are adjacent to the field of style transfer, and show great promise, we are not aware of diffusion-based methods which simultaneously offers the content-preservation, exemplar based style control, and feed forward inference speed of NST methods.

\section{BBST-4M Dataset}

We compile a large scale dataset of content and style images, using images from Behance.net for the style subset, and images from Flickr for the content subset. We filter based on the criteria that images must be high resolution (at least 1024px on the smallest side), and they must be stylistic for the style subset, and not stylistic for the content subset.

To filter based on the stylistic properties of the image, we first build a model to predict if an image is stylistic. We use ViT \cite{vit} for a content feature branch, and ALADIN-ViT \cite{stylebabel} for a style feature branch. We merge these two pre-trained branches with a small MLP for a final score. The model was trained iteratively, human-in-the-loop, on data from the StyleBabel dataset \cite{stylebabel}. We manually annotated images as being stylistic or not, with assistance from the model to help guide the process, with iterative re-training of the model with each pass of further dataset expansion through annotation. By the time the whole StyleBabel dataset was labelled, we found the model to be nearly perfect in predicting if an image was stylistic or not.

Our criteria to judge if an image is stylistic or not is subjective, but our general rule was that if it made sense to use an image as a target style, it was deemed stylistic. Images such as ordinary photographs or interface screenshots would be labelled as negative examples; while images such as oil paintings or digital art would be positive examples.

We used this stylistic detection model to filter out stylistic images from the Flickr data, and filter out non-stylistic images from the Behance data. The final dataset size is 2.2 million stylistic images from Behance, and 2 million content images from Flickr. 

\begin{figure*}
    \begin{center}
    \includegraphics[width=0.95\linewidth]{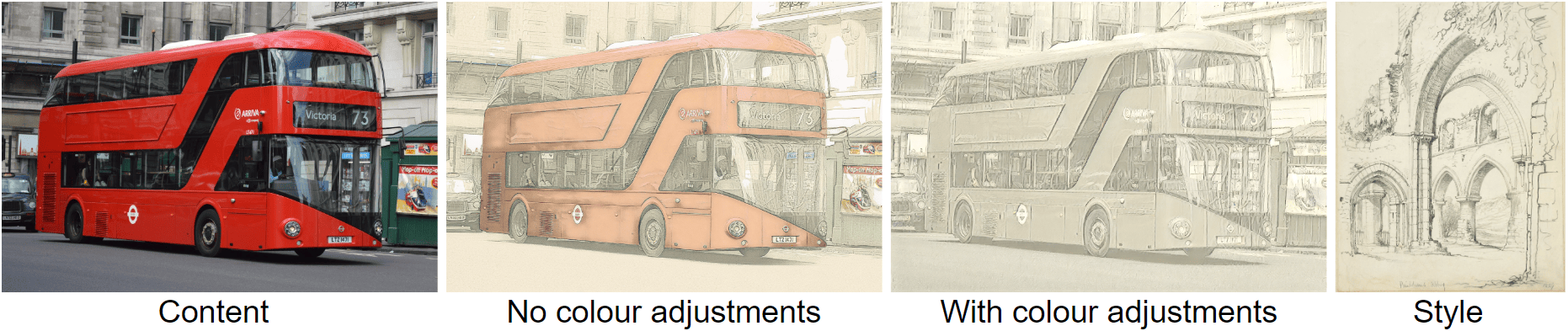}  
    \end{center}
    \squeezeup
    \squeezeup
   \caption{Visualization of the content colour bleeding problem, in content priors use 
   }
   \squeezeup
    \label{fig:prerecolour}
\end{figure*}

\section{Design decisions}

Works in the hypernetworks literature \cite{hyperstyle,hypernst} find that reformulating a model's task as the modification of the target domain can be more effective than generating completely new values. In the hypernetworks, this is done by generating weight deltas, to modify existing trained weights of a target model. This idea of generating deltas can similarly be applied to other tasks, such as style transfer. In our work, we propose this by designing our model to generate RGB deltas, over the existing content prior image. Generating a completely new image is unnecessary, as we already know what the structure of the image should be. For neural stylization, we are only interested in modifying the existing content image. So far, content preservation has been an area of focus for balancing design decisions in previous literature, but we propose this approach to alleviate this problem, as the content image already has all the needed information. An ablation vizualizing the benefit of our deltas-based approach can be seen in Fig \ref{fig:deltasAblation}.

There are nevertheless some challenges to overcome with this approach. First, a balance must be found in deciding how much information from the content image should be propagated. A stylized image should preserve the content of the reference image, but not necessarily retain all the fine details. For example, a stylized landscape image should rarely replicate details as fine as grass blades. We therefore first apply some light gaussian blurring on the content image, and apply bilateral filtering to further simplify the source image while preserving sharp edges. 
Finally, we first apply a weighting (0.5) to the content prior, to force the model to not rely purely on all the content prior's details (similar to how an artist traces over a faded version of the reference image). During inference, the blur strength can be adjusted/disabled, to allow slightly more or less original details to come through into the final image.

The second issue is colour information from the content image bleeding into the stylized image. We are only interested in the structure information, not the colours. There are several ways this issue can be mitigated, such as only using grayscale content priors, or using only priors for \textit{structure} channels in a different colour space such as HSL or OKLAB. Still, we find the best approach to be to re-colour the content prior with the style image colours. We adjust the mean and covariance, as per the work in \textit{Artistic Radiance Fields} \cite{art_rad_fields}. Figure \ref{fig:prerecolour} shows a visualization of the colour bleed from the content image when a content image is used as a prior, and how our pre-recolouring pipeline remedies it.

\subsection{Architecture and losses}
 
We first extract image features from the ground truth images using a pre-trained VGG \cite{vgg} model. Next, we use an attention based mechanism to induce the style transfer in feature space. We use 4 PAMA \cite{pama} blocks, between VGG features of \textit{style} and \textit{content} images. Finally, we decode these features using a decoder module. We use a domain-level adversarial loss \cite{contraAST} to support realism in the generated images, and identity losses to ensure strong information propagation.  
During the stylization process, we use contrastive losses amongst the generated stylized images to ensure consistency between the style and content. We further perform contrastive losses between the stylized images and the ground truth style images \cite{cast}. We use two crops of ground truth style for this, to raise the number of comparisons made to the style image whose style we replicate.

\begin{figure*}
    \begin{center}
    \includegraphics[width=0.9\linewidth]{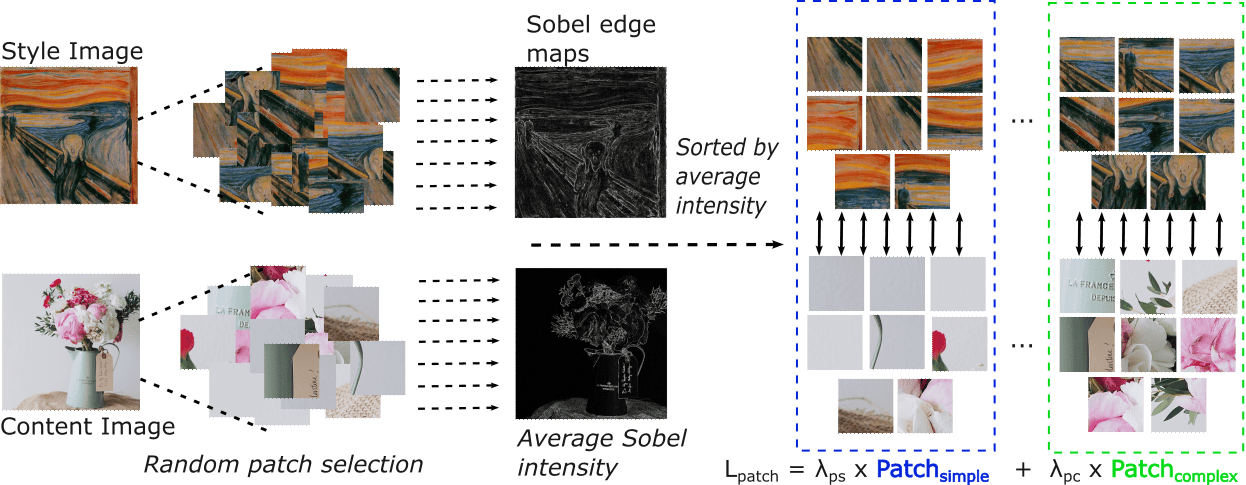}    
    \end{center}
   \squeezeup
   \caption{A visualization of the patch co-occurrence loss, specifically showing the Sobel guided selection process (Sec \ref{sec:style_halos}). Patches are randomly selected from both the stylized image, and the style image. Sobel edge maps of the content image and style image are used to compute average intensity scores for all patches, which are then sorted by this intensity score. Two patch co-occurrence losses are computed separately, for the simple patches, and the complex patches.}
    \label{fig:arch_sobel}
\end{figure*}

Finally, we use a patch co-occurrence discriminator \cite{sae} to provide an additional learning signal for style, via a patch based comparison. This further helps with small details in textures such as paint strokes, highly visible when generating high resolution images. This improves local style information, in addition to the global style information already supported by the other losses. We use a custom implementation of a patch co-occurrence discriminator, where we perform a more informed method of patch sampling via sobel edge maps (Sec. \ref{sec:style_halos}), to solve \textit{style halos}.

\subsection{Style Halos}
\label{sec:style_halos}

\begin{figure*}
    \begin{center}
        \includegraphics[width=0.9\linewidth]{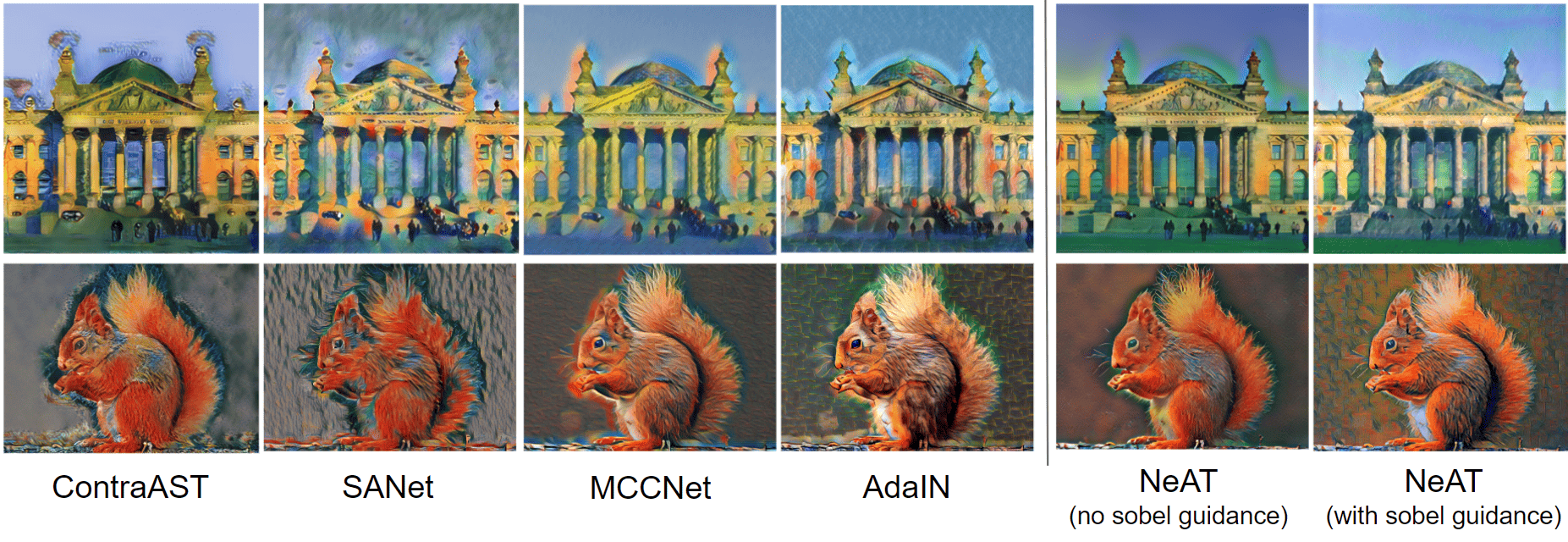}  
    \end{center}
    \squeezeup
    \squeezeup
    \squeezeup
    \caption{Visualization of the style halos intermittently present in many models. Style halos can be seen in the top row mostly around the two towers, in the sky. In the bottom row, they can be seen around the edges of the squirrel.}
    \label{fig:halos}
    \squeezeup
\end{figure*}

An issue we commonly observe in literature is the appearance of \textit{halos} in the generated style images. Visualized in Fig \ref{fig:halos}, these are outlines around objects which reduce the overall quality of the image. We notice that they mostly occur in low frequency areas adjacent to high frequency areas. The style halos themselves contain low frequency style textures, which are  appropriate for a low frequency area. The problem, therefore, is not with the areas inside the halos, but rather in the remaining low frequency region, further away from the edge. 
While previous work can sufficiently stylize transitions from high-frequency to low-frequency regions, they fail in entirely low-frequency regions. When relative edge information is not present locally, there is no guidance for how complex the texture/style needs to be. As a result, style complexity increases in low-frequency areas further from the boundaries, creating these \textit{halo}-like artifacts.

We propose a solution to this problem based on an informed selection of patches used in a patch co-occurrence discriminator, to provide a localized, non-global style signal. The vanilla discriminator selects 8 patches at random from the style image, and 8 at random from the generated image. Informed by our observations regarding \textit{style halos}, we improve this by splitting the loss into two separate patch co-occurrence discriminators, one acting on low frequency areas, and one acting on high frequency areas (Figure \ref{fig:arch_sobel}). 

We use Sobel edge maps to determine regions' frequency intensities for randomly sampled patches from style and content images. We then sample these regions from the style and stylized images and sort them by average intensity respective to their Sobel versions. The high average intensity patches are used in the high-frequency discriminator, and the low average intensity patches are used in the low-frequency discriminator. By forcing the discriminators to focus on high/low complexity patches separately, we provide a local signal that low-frequency regions of the content image should remain lower frequency \textit{relative} to the rest of the image when stylized. Thus, low-frequency regions far from boundaries are not stylized with overly complex texture for lack of a local relative signal, alleviating the halo artifacts.
The two losses are weighted by $\lambda_{ps}$ for the simple patches' loss and $\lambda_{pc}$ for the complex patches' loss. 
 
The full loss definition can be found in the supplementary materials. The final objective is shown in Eq 1, where $\mathcal{L}_{\text {s-contra }}$, and $\mathcal{L}_{\text {c-contra }}$ are standard contrastive losses. The $\mathcal{L}_{\text {patch}}$ terms are shown in Eq. 2 with \textit{SM}, \textit{sc}, \textit{s}, \textit{S}, \textit{D}, $\mathcal{C}$, and \textit{I} representing Sobel Maps, stylized image, style image, style dataset, discriminator, cropping function, and output image.
\squeezeup
\squeezeup


\begin{equation}
\hspace{-0.1cm}\mathcal{L}_{\text {final }}: = \lambda_1 \mathcal{L}_s+\lambda_2 \mathcal{L}_{\text {adv }}+\lambda_3 \mathcal{L}_c+\lambda_4 \mathcal{L}_{\text {identity }}+\lambda_5 \mathcal{L}_{\text {s-contra }}... 
\nonumber \squeezeup
\end{equation}
\begin{equation}
\hspace{0.3cm} +  \lambda_6 \mathcal{L}_{\text {c-contra }} + \lambda_7 \mathcal{L}_{\text {patch\_simple}} + \lambda_8 \mathcal{L}_{\text {patch\_complex}} 
\hspace{0.5cm}\label{eq:final_objective}
\end{equation}

\begin{equation}
\hspace{-0.3cm}\mathcal{L}_{\text {patch }} = \\ \underset{I_s \sim S}{\mathbb{E}}[-\log (D_{\text {patch }} (\mathcal{C} ( I_{s c}, SM_{s c} ), \mathcal{C}(I_{s}, SM_{s} ) ) ) ]
\label{eq:patch}
\end{equation}
\squeezeup
\squeezeup
\squeezeup

\section{Experiments}
\label{sec:experiments}

We can perform interpolation in the style space, by interpolating between the feature maps fed through the decoder. 
Fig 5 in the supplementary 
visualizes interpolation between the feature map of a content image, compiled for reconstruction, and the feature map of the same content image, stylized by the style image shown. For full interpolation, we additionally interpolate the content image prior due to our pre-processing pipeline. We use an alpha $\alpha$ value to signify the interpolation strength between these. 

Interestingly, we can \textit{amplify} the stylization strength by increasing $\alpha$ beyond 1, thus further boosting the effect of the stylization. A couple such samples, denoted in 
the figure 
as $\alpha=1.25$ and $\alpha=1.5$ show how the stylized image's style even more strongly resembles that of the style image. Yet, structural details in the image are still maintained - such as the fence posts and the birds on the wire. Similar to the content prior's blur parameters, changing the alpha value during inference can be used by an artist as a tool to control the stylization process further.

\begin{figure}
    \begin{center}
    \includegraphics[width=1\linewidth]{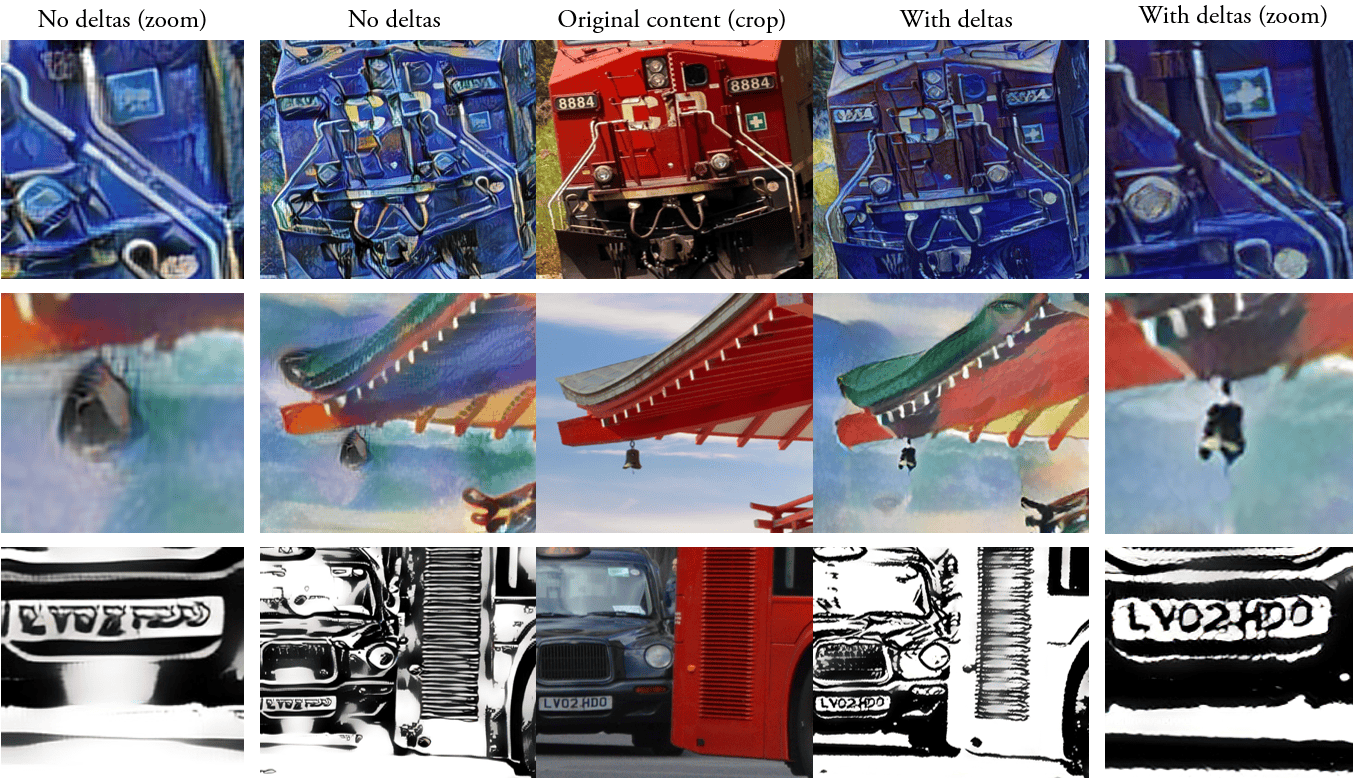}  
    \end{center}
    \squeezeup\squeezeup
    \caption{Ablation showing detail gain. (Left) RGB (Right) Deltas}
    \squeezeup
    \label{fig:deltasAblation}
\end{figure}

\subsection{Training and evaluation}

The strengths of the content prior blurs can be varied, but for training, we settle on a balanced amount of 7px blurring, followed by bilateral filtering with filter size 25, and a sigma value of 100.
During early tests, we experimented with unbalanced averaging of the losses from the two sobel guided patch co-occurrence discriminators. Based on the idea that high frequency information requires more work to learn correctly, we provide more of the loss budget to the high frequency patches, weighing them at \textit{0.75} ($\lambda_{pc}$), versus \textit{0.25} ($\lambda_{ps}$) for the low frequency patches. The training time for NeAT is roughly 3 days on an A100.

We evaluate four aspects of our model's generated images compared to literature, in automated metrics: (1) colour consistency in regards to the style image, (2) content structure preservation in regards to the content image, (3) style consistency compared to the style image, (4) inference run-time of the model to run the style transfer, end-to-end.

We measure colour consistency using the Chamfer distance. We use a version of this normalized by the number of pixels in the image, such that the score is independent of the image resolution. We use LPIPS \cite{lpips} to measure content preservation, and we calculate the average inference run-time, averaged over our test set. To measure style consistency between the style and stylized images, we use the SIFID metric introduced in SinGAN \cite{singan}. This metric measures FID score, but only amongst individual images. SIFID was used in previous works such as Swapping Autoencoders \cite{sae} for measuring photorealistic style transfer - we use it for measuring artistic style transfer. We capture these results in Tables \ref{tab:metrics_results} and \ref{tab:results}. We further include results from our model where the prior image blurring is turned off. This is a control vector that an artist can use for adjusting how many details from the content image should come through into the final, stylized image. Turning the blurring off can thus further improve the preservation of details, measured in the table via LPIPS.

Finally, we undertake a user study to measure stylization quality and consistency with the target image.

\begin{table}[h!]
      \centering
      \small
  
      \begin{tabular}{l|c|c|c}
        \hline
        Model & LPIPS & SIFID & Chamfer \\

        \hline
        ContraAST \cite{contraAST} & 0.643 & 2.519 & 51.696  \\ %
        PAMA \cite{pama} & 0.656  & 2.301 & 271.630  \\ %
        SANet \cite{sanet} & 0.684 & 2.837 & 58.297  \\ %
        CAST \cite{cast} & 0.579 & 1.729 & 214.075  \\ %
        AdaAttn \cite{adaattn} & 0.598 & 3.157 & 196.391  \\ %
        MCCNet \cite{mccnet} & 0.646 & 2.689 & 393.348  \\ %
        ArtFlow \cite{artflow} & 0.346 & 4.646 & 656.336  \\ %
        NNST \cite{nnst} & 0.626  & 3.784 & 2617.187  \\ %
        AdaIN \cite{adain} & 0.666 & 2.235 & 204.790 \\ %
        \hline
        \hline
        NeAT (Ours - WikiArt+MSCoco) & 0.655 & 1.171 & 78.780  \\ %
        $\longrightarrow$ \textit{no prior blurring}  & 0.596 & 2.343 & 96.541  \\ %
        \hline
        NeAT (Ours - BBST-4M) & 0.687 & 1.053 & 41.742  \\
        $\longrightarrow$ \textit{no prior blurring}  & 0.635 & 1.306 & 54.227  \\
        \hline
      \end{tabular}

    \caption{
    Quantitative evaluation metrics. Lower is better.
    }
    \label{tab:metrics_results}
  \squeezeup
  \end{table}

\begin{table}[h!]
  \centering
  \begin{adjustbox}{width=0.9\linewidth}
      \centering
      \small
  
      \begin{tabular}{l|c|c|c}
        \hline
        Model & 256x256 & 512x512 & 1920x1080 \\

        \hline
        ContraAST \cite{contraAST} & 0.030 & 0.043 & 0.150 \\ %
        PAMA \cite{pama} & 0.035 & 0.051 & 0.189 \\ %
        SANet \cite{sanet} & 0.030 & 0.043 & 0.163 \\ %
        CAST \cite{cast} & 0.050 & 0.055 & 0.184 \\ %
        AdaAttn \cite{adaattn} & 0.053 & 0.068 & 0.221 \\ %
        MCCNet \cite{mccnet} & 0.040 & 0.054 & 0.178 \\ %
        ArtFlow \cite{artflow} & 0.153 & 0.247 & 1.178 \\ %
        NNST \cite{nnst} & 17.647 & 17.266 & 65.367 \\ %
        AdaIN \cite{adain} & 0.037 & 0.0517 & 0.185 \\ %
        \hline
        \hline
        NeAT (Ours) & 0.049 & 0.068 & 0.226 \\ %
        $\longrightarrow$ \textit{no prior blurring} & 0.047 & 0.058 & 0.196 \\ %
        \hline
      \end{tabular}
   \end{adjustbox}

    \caption{
    Timing (seconds/image) of all methods. Lower is better
    }
    \label{tab:results}
  \squeezeup
  \squeezeup
  \end{table}

\begin{table}[t!]
  
      \centering
      \small
  
      \begin{tabular}{l|c|c|c}
        \hline
        Model & Overall & Content & Style \\
        \hline
        
        ContraAST \cite{contraAST} & 62.5\% & 64.0\% & 66.5\% \\
        PAMA \cite{pama} & 63.5\% & 71.0\% & \underline{54.5\%} \\
        SANet \cite{sanet} & 78.5\% & 78.0\% & 64.0\% \\
        CAST \cite{cast} & 60.0\% & 57.0\% & 69.0\% \\
        AdaAttn \cite{adaattn} & \underline{52.5\%} & 54.5\% & 56.0\% \\
        MCCNet \cite{mccnet} & 58.5\% & 55.5\% & 66.0\% \\
        ArtFlow \cite{artflow} & 60.0\% & \underline{51.5\%} & 72.0\% \\
        NNST \cite{nnst} & 84.5\% & 83.0\% & 81.5\% \\ 
        AdaIN \cite{adain} & 72.5\% & 74.5\% & 70.0\% \\
        
        \hline
        \textit{Average} & 65.83\% & 65.44\% & 66.61\% \\
        \hline
      \end{tabular}

    \caption{
    Preference of our method on AMT, compared to baselines (61\% agreement). A value higher than $50\%$ indicates workers preferred our method - we beat all baselines, on each of the 3 separate experiments. We underline the closest baseline to NeAT.
    }
    \label{tab:results_mturk}
    \squeezeup
    \squeezeup
\end{table}

\begin{figure*}
    \begin{center}
    \includegraphics[width=0.82\linewidth]{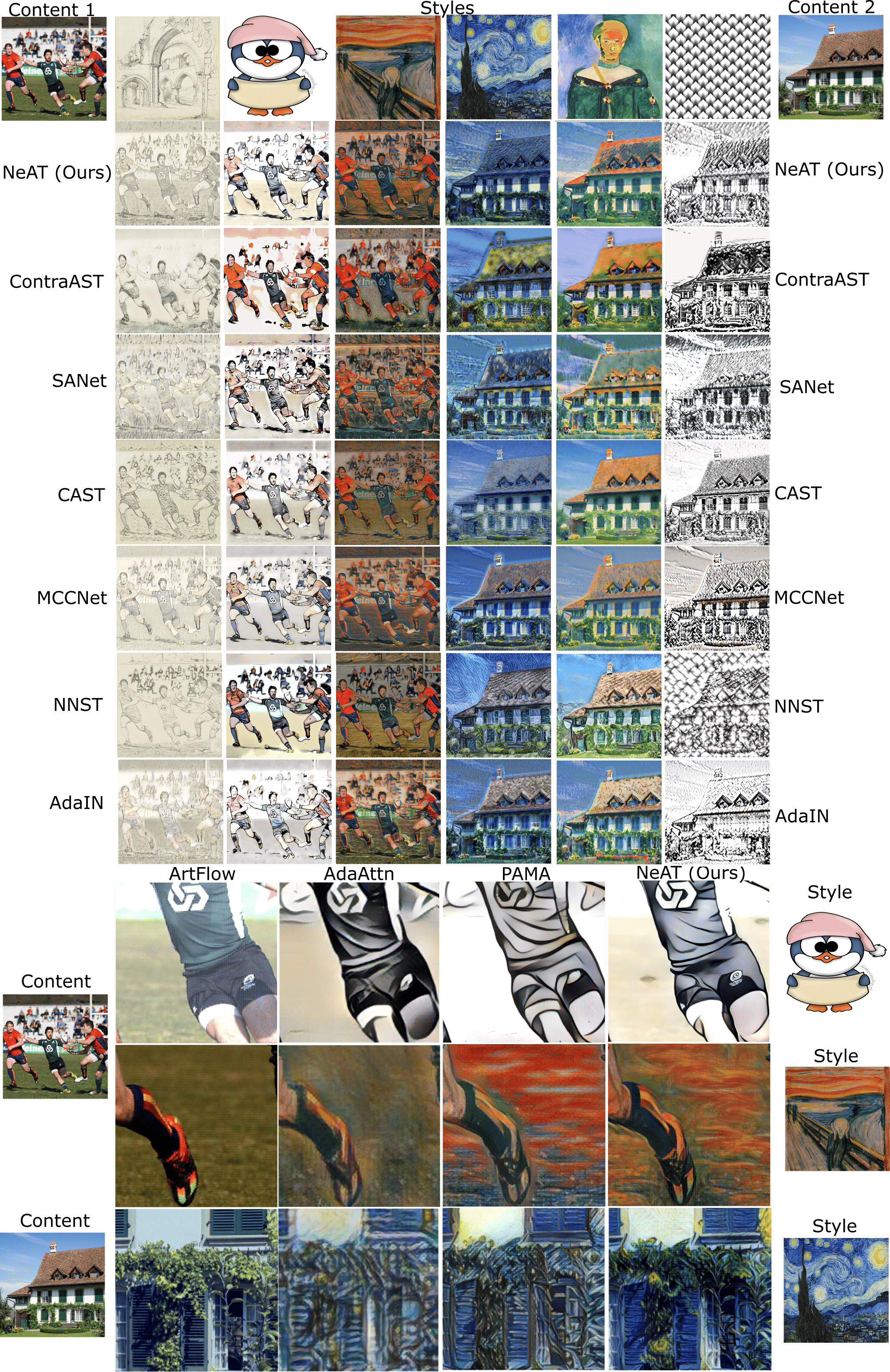}  
    \end{center}
    \squeezeup
    \squeezeup
   \caption{Visualization of style transfer using NeAT, and the baseline methods. Please zoom for more details.}

    \label{fig:baselines_viz}
\end{figure*}

\subsection{User study}

We conduct a user study on Amazon Mechanical Turk (AMT), to measure real life preference between our model, and baseline models. We show users the content and style image in the middle of the screen, and on either side, a stylized sample using either our model, or a baseline comparison model. The whole sample is shown at the top, and a centre crop shown below, to highlight smaller details in the image - critical when working with high resolution images. The left/right columns for either models is randomly shuffled. We test our model against the baselines, across three different experiments: \textit{Overall}, \textit{Content}, and \textit{Style}. All three use NeAT trained on WikiArt/MSCoco, for a fair comparison against literature.

The \textit{Content} experiment asks users to evaluate the semantic structure and detail preservation between the content and stylized images. The \textit{Style} experiment asks users to assess the similarity of artistic style between the output and target style, irrespective of content. The \textit{Overall} experiment asks users to evaluate both criteria simultaneously to gauge the overall combined quality of the style transfer.

We form 750 test images from 15 content images and 50 style images.
From this, we randomly sub-sample 200 stylized pairs for each experiment. We ask 5 different workers to pick their preference for each pair, and we use consensus voting to determine the overall decision for each pair. In total, 63 total workers are contributing to the user study. We collect the results in Table \ref{tab:results_mturk}. Finally, we qualitatively visualize random stylization results across NeAT and the baselines in Figure \ref{fig:baselines_viz}. We enlarge the visualizations from the closest baselines in the user study.

\subsection{Generalization}
\label{sec:generalizability}
We use BBST-4M, our newly collected dataset, to improve style transfer quality, especially across styles not covered in typical style datasets such as WikiArt. Such datasets are limited in their scope to mainly fine art images. In contrast, the collection process for the style subset of BBST-4M covered a much larger variety of styles, as typically found on \textit{Behance.net}. We train our model using BBST-4M, and we evaluate via two further user studies how the quality of the style transfer changes. As before, we first test the BBST-4M variant with the test set. Second, we separately evaluate how well the model generalizes to non fine art images, by evaluating using an \textit{out of distribution} set of style images, respective to WikiArt. We build this small set of images from a separate data scrape from Behance, by selecting images that are, on average, far away in ALADIN style space relative to the images in the WikiArt dataset. These are mainly digital art, vector art, and other such images not typically seen in such fine art collections.

As before, users on AMT evaluate stylized pairs of the same content and style image for two variants. The user studies indicate a preference for the BBST-4M variant at $56.0\%$ over the WikiArt variant. Similarly, there is a preference of $60.0\%$ when using the \textit{out of distribution} style test set. This more significant gap indicates that the model has more generalization capabilities than the WikiArt model.

\section{Limitations}

\begin{figure}
    \begin{center}
    \includegraphics[width=1\linewidth]{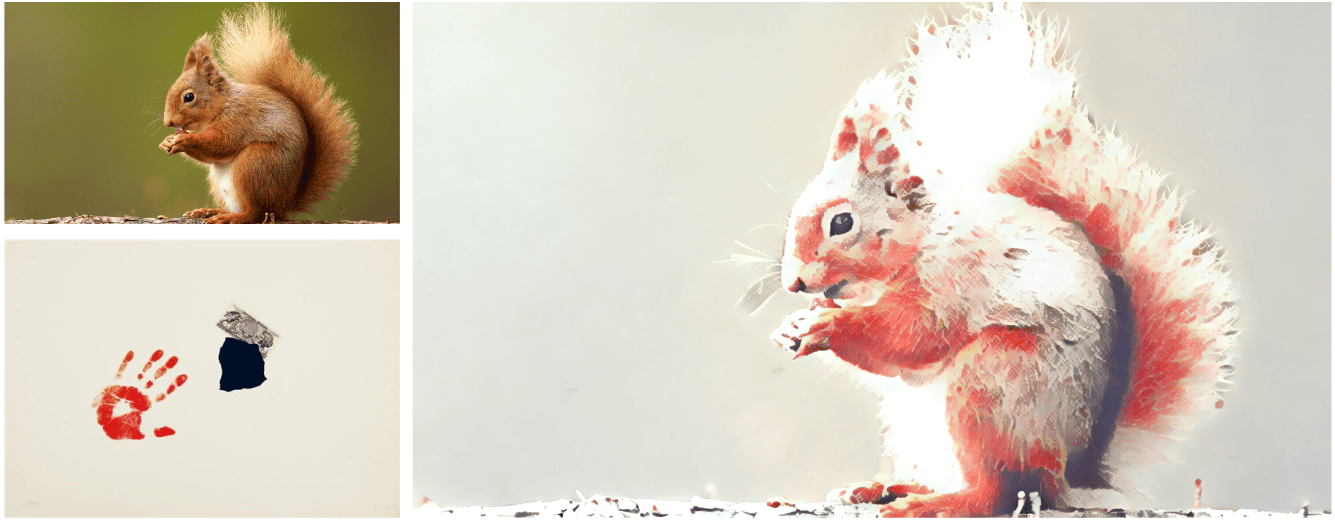}  
    \end{center}
    \squeezeup\squeezeup
    \caption{Example of limitation on color adjustments. The squirrel's tail is overexposed, and details are lost.}
    \squeezeup
    \squeezeup
    \label{fig:limitation}
\end{figure}

We use simple adjustments to the mean and covariance for the pre-recoloring of the content image prior. While effective, this process could be better and can lead to occasional inconsistencies and artifacts in the image colors. We visualize this example in Figure \ref{fig:limitation}, where imperfections in the color adjustments can lead to lost details.

Finally, though not an issue with our method precisely, different resolutions of the content image lead to noticeably different stylized outputs. An artist can remedy this by attempting the process with multiple resolutions. However, this is an additional \textit{hyper-parameter} that needs to be explored during inference. Though it could be considered a positive, as it allows control over the level of style complexity in the edited image.

\section{Conclusion}
\label{sec:conclusion}

We presented NeAT, a new style transfer model, achieving state-of-the-art performance through novel technical design decisions. We stylize images through editing rather than regeneration, leading to better content preservation and improved artistic style transfer. We provide a solution to style halos appearing in stylized images, through careful guidance during patch selection for a patch co-occurrence loss. We use Sobel edge maps to sort and separate high frequency and low frequency areas, thereby better matching appropriate areas of the style image to match regions in the content image for the stylization.

We additionally propose and release a model that can predict how stylistic an image is. We use this model to pre-process the BBST-4M dataset we also release, the first large scale high resolution style transfer dataset.

{\small
\bibliographystyle{ieee_fullname}
\bibliography{egbib}
}

\appendix

\twocolumn[{%
\renewcommand\twocolumn[1][]{#1}%
\maketitle
\vspace{-30pt}

\begin{center}
    \centering
    \includegraphics[width=1\linewidth]{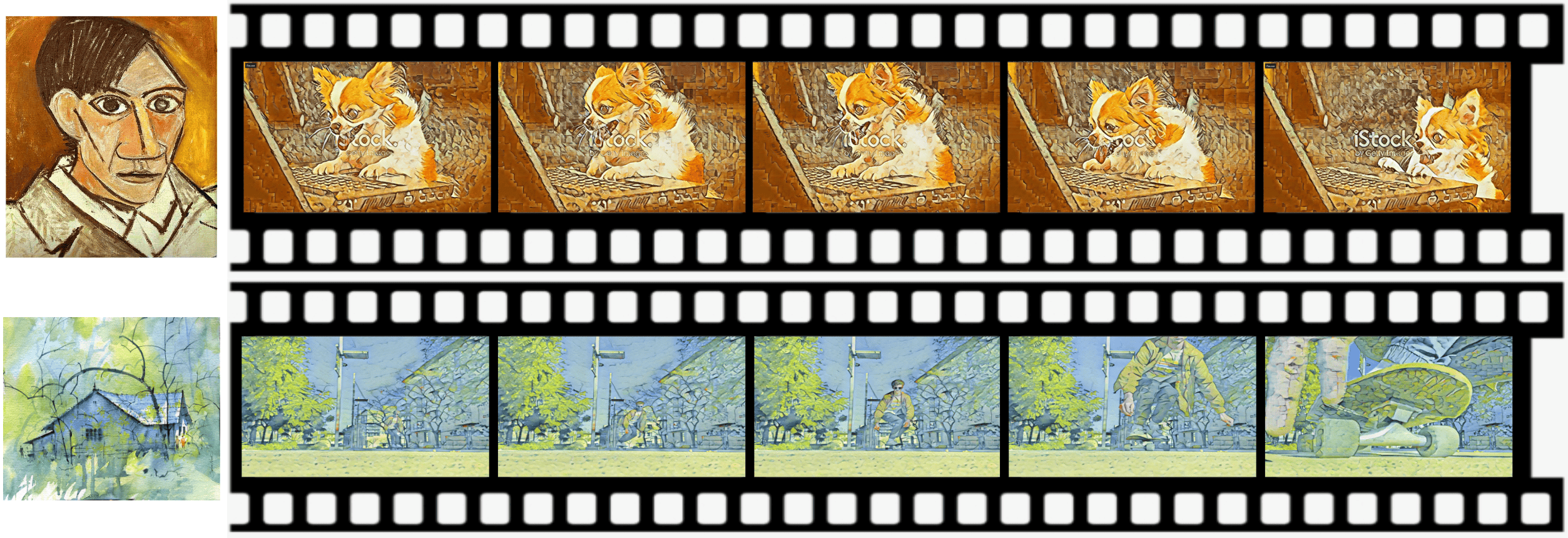}
    \label{fig:film}
Figure 1. Video style transfer using NeAT
\end{center}%
\vspace{9pt}
}
]
\setcounter{figure}{1}

\begin{figure*}[htp!]
    \centering
    \includegraphics[width=1\linewidth]{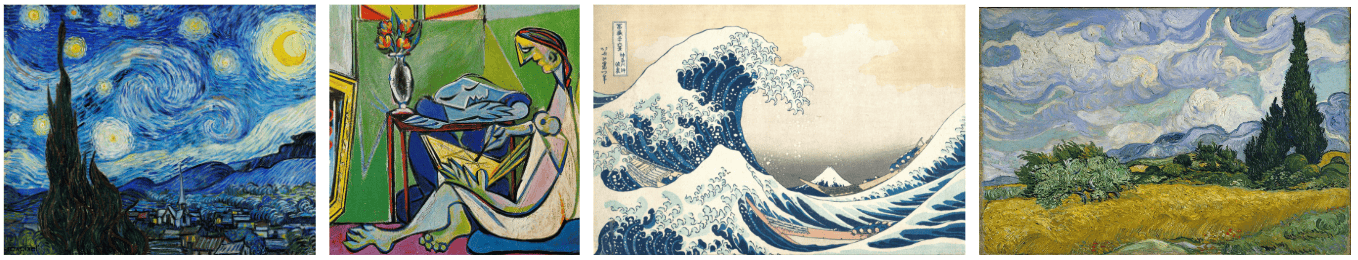} 
    \caption{Example \textit{fine art} style images used in the test set}
    \label{fig:fineart_style}
\end{figure*}

\begin{figure*}[htp!]
    \centering
    \includegraphics[width=1\linewidth]{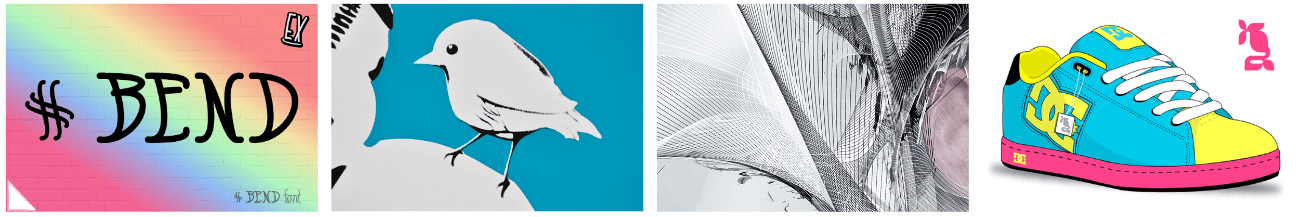} 
    \caption{Example \textit{out of WikiArt distribution} style images used in the test set}
    \label{fig:ood_style}
\end{figure*}

\section{Example BBST-4M images}

\begin{figure*}[htp!]
    \centering
    \includegraphics[width=1\linewidth]{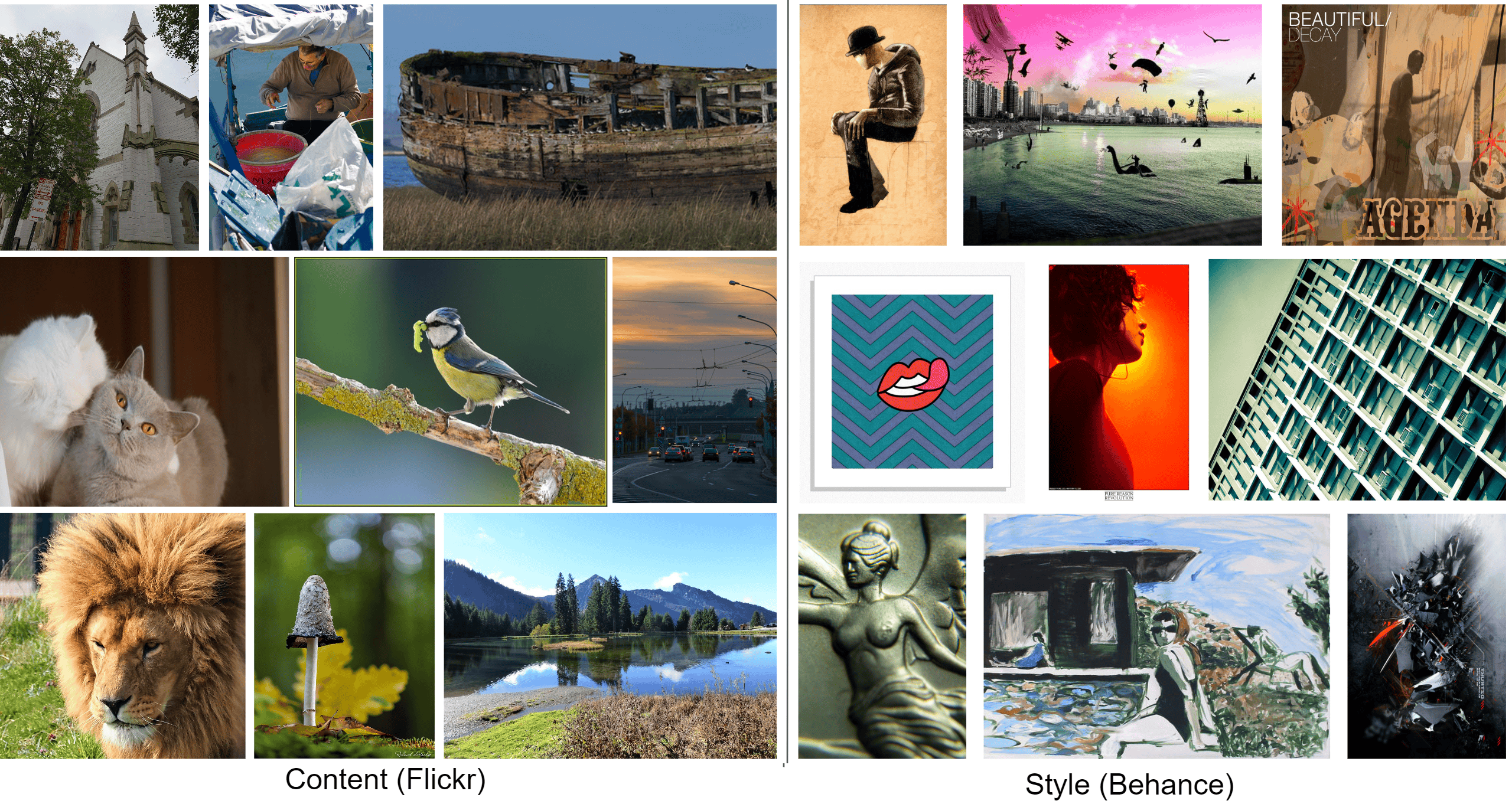} 
    \caption{BBST-4M data examples from the content and style splits}
    \label{fig:bbst_samples}
\end{figure*}

We propose and release BBST-4M, a novel dataset focusing on large scale subsets of \textit{content} and \textit{style} with high resolution images. We collect the \textit{content} images from \textit{Flickr}, and we filtered out stylistic images using our stylistic detection model. We build our \textit{style} images using data from \textit{Behance.net}. We use this platform as it contains a vast diversity of styles, as seen in previous datasets such as BAM, BAM-FG, and StyleBabel - all collected from Behance. One of our aims for NeAT has been strong generalization capabilities. We measured and achieved this using this much wider range of styles, found in the style set of BBST-4M. Figure \ref{fig:ood_style} shows some differences compared to WikiArt, but we also include some random example images from both the \textit{content} and \textit{style} sets in Figure \ref{fig:bbst_samples}.

\section{Example test set images}

We perform two separate user studies to evaluate the performance of our BBST-4M dataset against the standard WikiArt+MSCoco datasets. We separately evaluate how well it performs on more traditional \textit{fine art} style images, as typically found in WikiArt using fine art style images. To measure generalization capabilities, we use an \textit{out of distribution} set of images, relative to the type of fine art styles found in WikiArt. Figure \ref{fig:fineart_style} visualizes fine art images, typical in WikiArt. Figure \ref{fig:ood_style} visualizes images that are unlike the kind of images present in WikiArt. These are pulled separately from Behance.net, and were deemed dissimilar to WikiArt through being, on average, far away in ALADIN style space to the fine art images of WikiArt.

\section{Further examples of style transfer using NeAT}

We present further style transfer examples created using BBST-4M NeAT in Figures \ref{fig:extra1}, \ref{fig:extra2}, and \ref{fig:extra3}. We encourage readers to zoom in, for viewing details.

\section{Experimenting with video stylization}

As an additional downstream use case, we explore the use of NeAT for video stylization. We explore a simple approach of splitting a video into frames using \textit{ffmpeg}, performing style transfer on every frame, and stitching the stylized frames back together into a video again using \textit{ffmpeg}. As a simple initial experiment, this does not incorporate any explicit temporal consistency designs, but as we show in the included videos, NeAT serves as a promising backbone for future work that could incorporate further design modules such as \textit{EbSynth}. Our use of content priors help to stabilise the frame-to-frame continuity of style transfer. Figure 1 shows a brief visualization of two video clips, stylized with NeAT. 

\onecolumn
\section{Further details on loss objective}


The full expanded mathematical definition of our loss objective is detailed in this section. Eq \ref{eq:Ls} details $\mathcal{L}_s$, the style loss computed amongst style and stylized image features, where $\phi_i$ represents VGG-19 layer index, with $\mu$ and $\sigma$ representing mean and standard deviation of extracted feature maps. $I_s$ represents style image from the style dataset $S$, $I_c$ represents content image from the content dataset $C$, and $I_sc$ represents the final stylized image, after applying the generated RGB deltas over the original $I_c$.

\vspace{-1cm}
\begin{equation}
    \mathcal{L}_s:=\sum_{i=1}^L\left\|\mu\left(\phi_i\left(I_{s c}\right)\right)-\mu\left(\phi_i\left(I_s\right)\right)\right\|_2+\left\|\sigma\left(\phi_i\left(I_{s c}\right)\right)-     
    \sigma\left(\phi_i\left(I_s\right)\right)\right\|_2
    \label{eq:Ls}
\end{equation}

\vspace{-1cm}
Eq \ref{eq:Ladv} represents the domain-level adversarial loss, as per \cite{contraAST}, learning to discriminate between generated stylized images, and real artworks. Here, a discriminator $\mathcal{D}$ operates over the stylized image, following encoder $E$, transformation $T$, and decoder $D$ modules.

\vspace{-1cm}
\begin{equation}
    \mathcal{L}_{a d v}:=\underset{I_s \sim S}{\mathbb{E}}\left[\log \left(\mathcal{D}\left(I_s\right)\right)\right]+ \underset{I_c \sim C, I_s \sim S}{\mathbb{E}}\left[\log \left(1-\mathcal{D}\left(D\left(T\left(E\left(I_c\right), E\left(I_s\right)\right)\right)\right)\right)\right]
    \label{eq:Ladv}
\end{equation}

\vspace{-2cm}
Eq \ref{eq:Lc} details standard perceptual loss, and Eq \ref{eq:Lidentity} details an identity loss, to preserve the same image when the content and style images are the same.

\vspace{-2cm}
\begin{equation}
    \mathcal{L}_c:=\left\|\phi_{\text {conv4\_2 } }\left(I_{s c}\right)-\phi_{\text {conv4\_2}}\left(I_c\right)\right\|_2
    \label{eq:Lc}
\end{equation}
\vspace{-2cm}
\begin{equation}
    \mathcal{L}_{\text {identity }}:=\lambda_{\text {identity } 1}\left(\left\|I_{c c}-I_c\right\|_2+\left\|I_{s s}-I_s\right\|_2\right)+ \lambda_{\text {identity } 2} \sum_{i=1}^L\left(\left\|\phi_i\left(I_{c c}\right)-\phi_i\left(I_c\right)\right\|_2+\left\|\phi_i\left(I_{s s}\right)-\phi_i\left(I_s\right)\right\|_2\right)
    \label{eq:Lidentity}
\end{equation}

Eqs \ref{eq:LcontraS} and \ref{eq:LcontraC} show contrastive losses as detailed in Sec 4.1, similar to \cite{contraAST} and \cite{cast}, where \textit{h} is a projection head $l_s$ and $l_c$ are style/content embedding extraction respectively, and $\tau$ is the temperature hyper-parameter.
\begin{equation}
    \mathcal{L}_{s-\text { contra }}:=-\log \left(\frac{\exp \left(l_s\left(s_i c_j\right)^T l_s\left(s_i c_x\right) / \tau\right)}{\exp \left(l_s\left(s_i c_j\right)^T l_s\left(s_i c_x\right) / \tau\right)+\sum \exp \left(l_s\left(s_i c_j\right)^T l_s\left(s_m c_n\right) / \tau\right)}\right)
    \label{eq:LcontraS}
\end{equation}
\begin{equation}
    \mathcal{L}_{c-\text { contra }}:=-\log \left(\frac{\exp \left(l_c\left(s_i c_j\right)^T l_c\left(s_y c_j\right) / \tau\right)}{\exp \left(l_c\left(s_i c_j\right)^T l_c\left(s_y c_j\right) / \tau\right)+\sum \exp \left(l_c\left(s_i c_j\right)^T l_c\left(s_m c_n\right) / \tau\right)}\right)
    \label{eq:LcontraC}
\end{equation}

The $\mathcal{L}_{\text {patch }}$ term defined in Eq \ref{eq:Lpatch} is our patch discriminator $(D_{\text {patch }}$ loss, guided by Sobel Maps ($SM$).
\begin{equation}
    \mathcal{L}_{\text {patch }} = \\ \underset{I_s \sim S}{\mathbb{E}}[-\log (D_{\text {patch }} (\operatorname{crop} ( I_{s c}, SM_{s c} ), \operatorname{crops}(I_{s}, SM_{s} ) ) ) ] 
    \label{eq:Lpatch}
\end{equation}

Our final combined loss objective is repeated in \ref{eq:Lfinal}.
\begin{equation}
    \mathcal{L}_{\text {final }}:=\lambda_1 \mathcal{L}_s+\lambda_2 \mathcal{L}_{\text {adv }}+\lambda_3 \mathcal{L}_c+\lambda_4 \mathcal{L}_{\text {identity }}+\lambda_5 \mathcal{L}_{\text {s-contra }}+\lambda_6 \mathcal{L}_{\text {c-contra }} + \lambda_7 \mathcal{L}_{\text {patch\_simple}} + \lambda_8 \mathcal{L}_{\text {patch\_complex}}
    \label{eq:Lfinal}
\end{equation}


\section{Interpolation strength}

\begin{figure}[h]
    \begin{center}
    \includegraphics[width=1\linewidth]{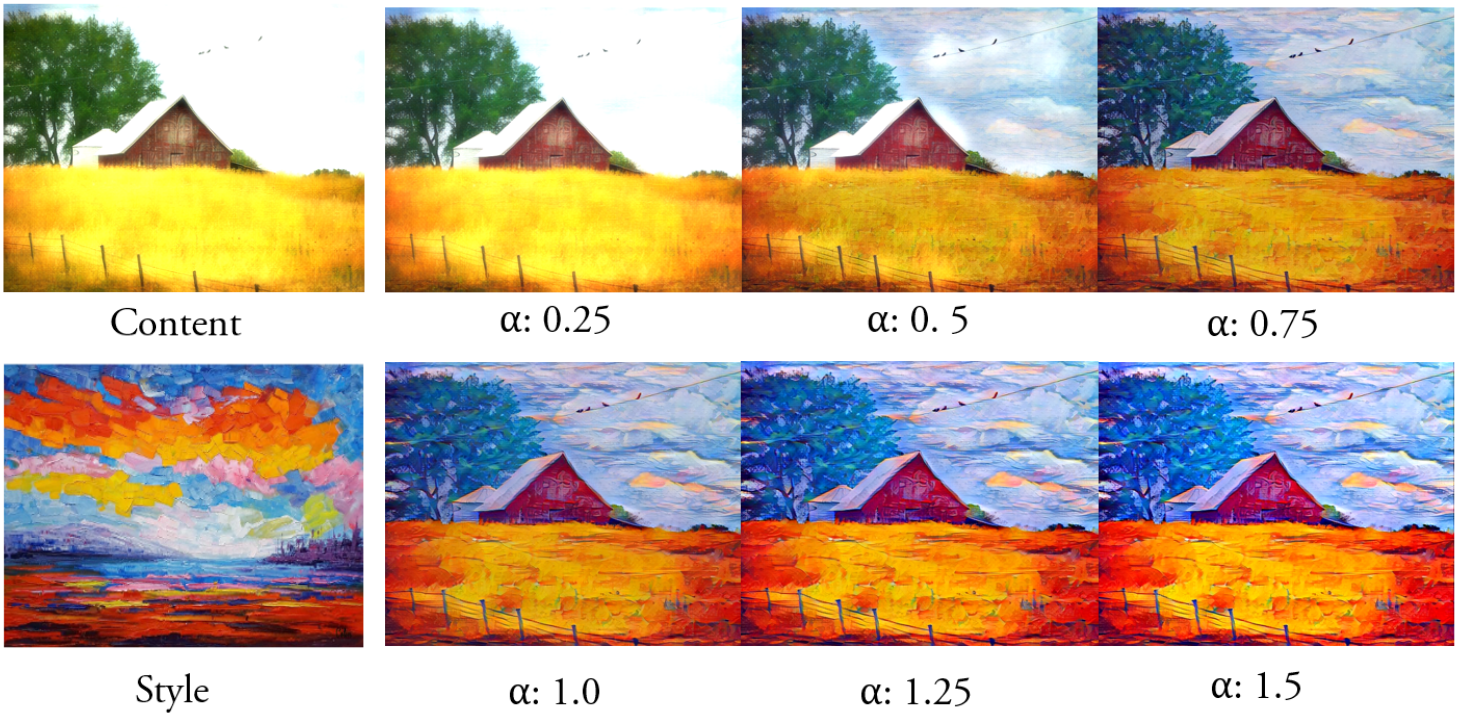}  
    \end{center}
    \squeezeup\squeezeup
   \caption{Interpolation between the original image and its stylized variant, by interpolating the features input to the decoder.}
    \label{fig:interp}
\end{figure}

Figure \ref{fig:interp} visualizes style interpolation and boosting.

\twocolumn

\begin{figure*}[htp!]
    \centering
    \squeezeup
    \squeezeup
    \includegraphics[width=0.75\linewidth]{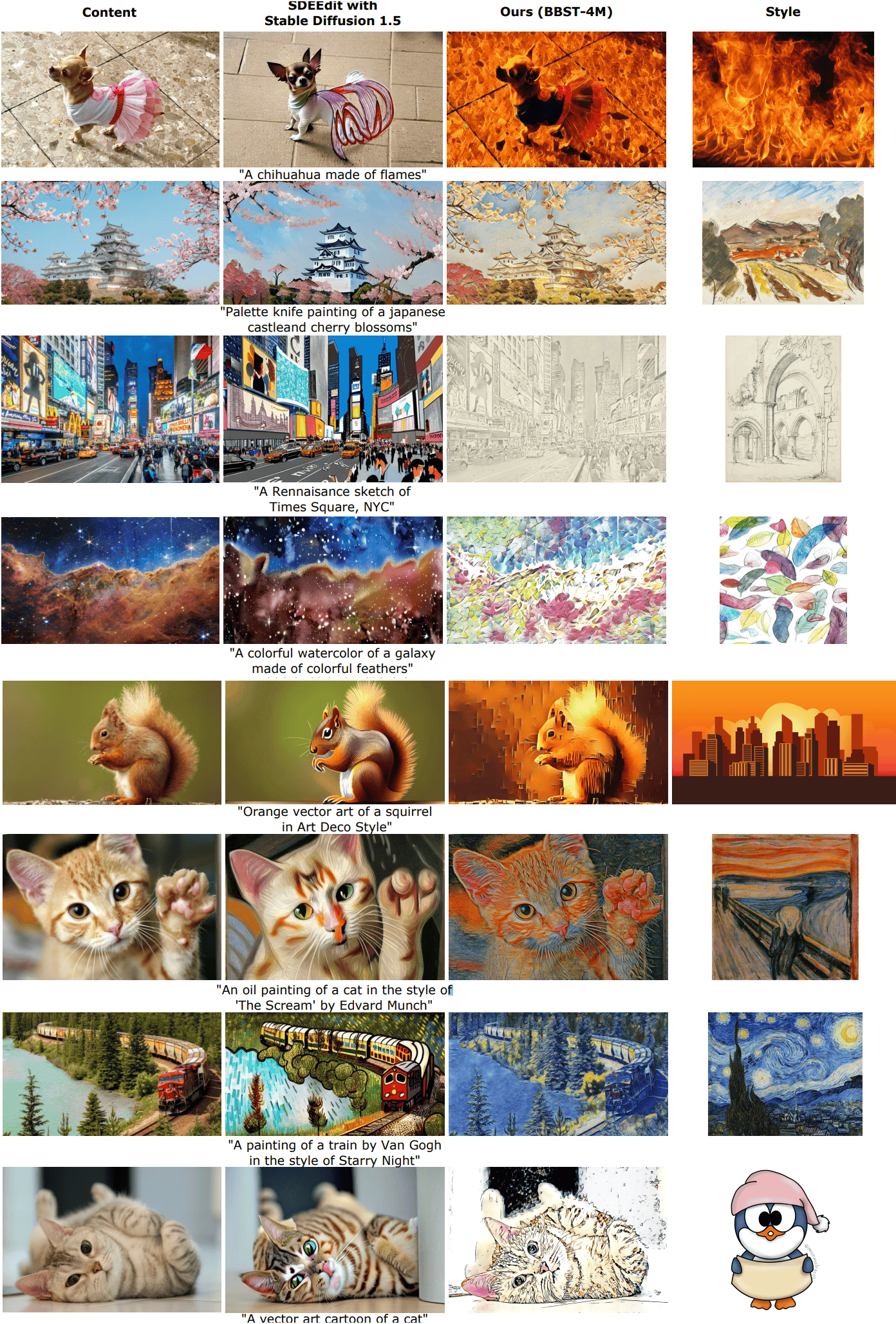} 
    \squeezeup
    \caption{Comparison between NeAT trained on BBST-4M (Ours) and SDEdit \cite{sdedit} using Stable Diffusion 1.5. We use $t_0=0.5$, 100 ddim steps, and $\eta=1.$ for all SDEdit results after sweeping these three hyperparameters for the overall best results (and fastest speed without degrading results). Captions used to guide SDEdit are under each Stable Diffusion result. Note that it is difficult to guide style transfer using language. Even in cases of famous paintings, like Starry Night, Stable Diffusion only matches the style of Van Gogh, not the particular painting. Furthermore it is difficult to consistently trade off between content preservation and stylization with fixed hyperparameters. While Stable Diffusion produces pleasing results for the Japanese castle, squirrel, and train, the other results are either under-stylized or have artifacts (e.g. the cats’ faces). While this comparison is imperfect, and there are many other diffusion techniques that may prove to be important components of a diffusion based stylization pipeline \cite{prompt2prompt,imagic,unitune,textualinversion,dreambooth}, we are not aware of any work that studies their combination in a systematic way for style transfer. We believe these results serve as a sensible qualitative comparison with diffusion based stylization in the absence of such work.}
\end{figure*}

\begin{figure*}[t!]
    \centering
    \includegraphics[width=0.95\linewidth]{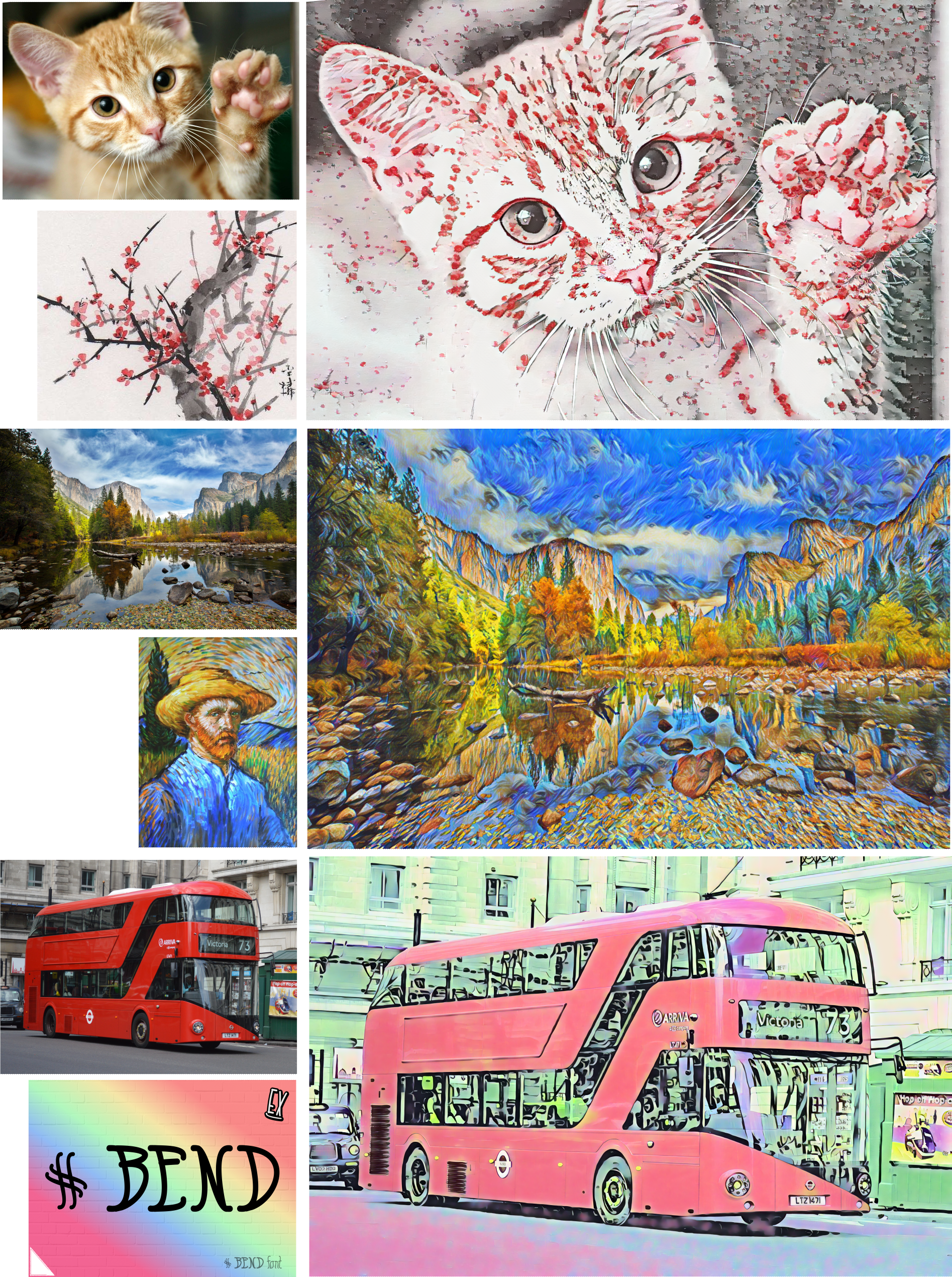} 
    \caption{}
    \label{fig:extra1}
\end{figure*}

\begin{figure*}[t!]
    \centering
    \includegraphics[width=0.95\linewidth]{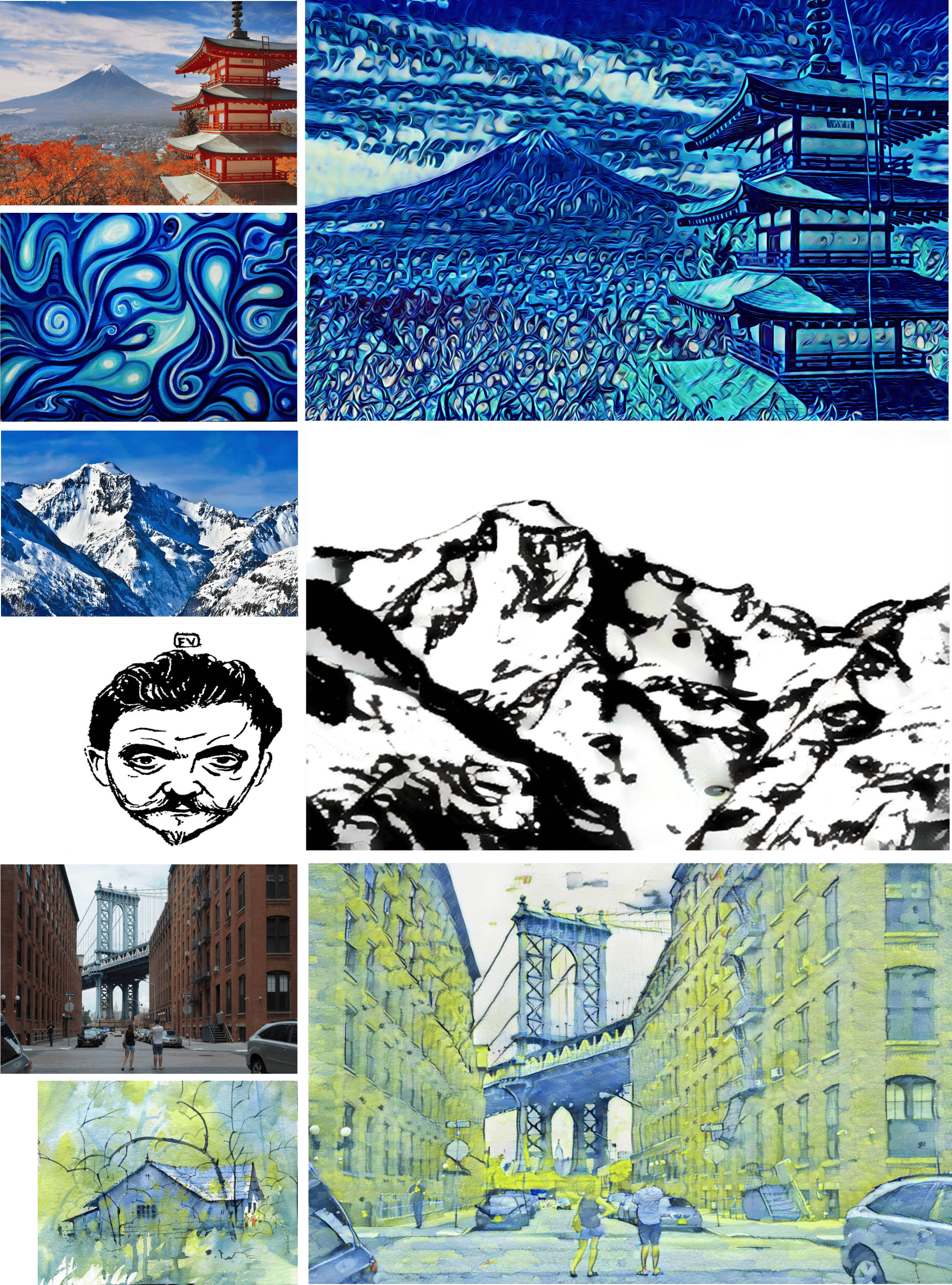} 
    \caption{}
    \label{fig:extra2}
\end{figure*}

\begin{figure*}[t!]
    \centering
    \includegraphics[width=0.95\linewidth]{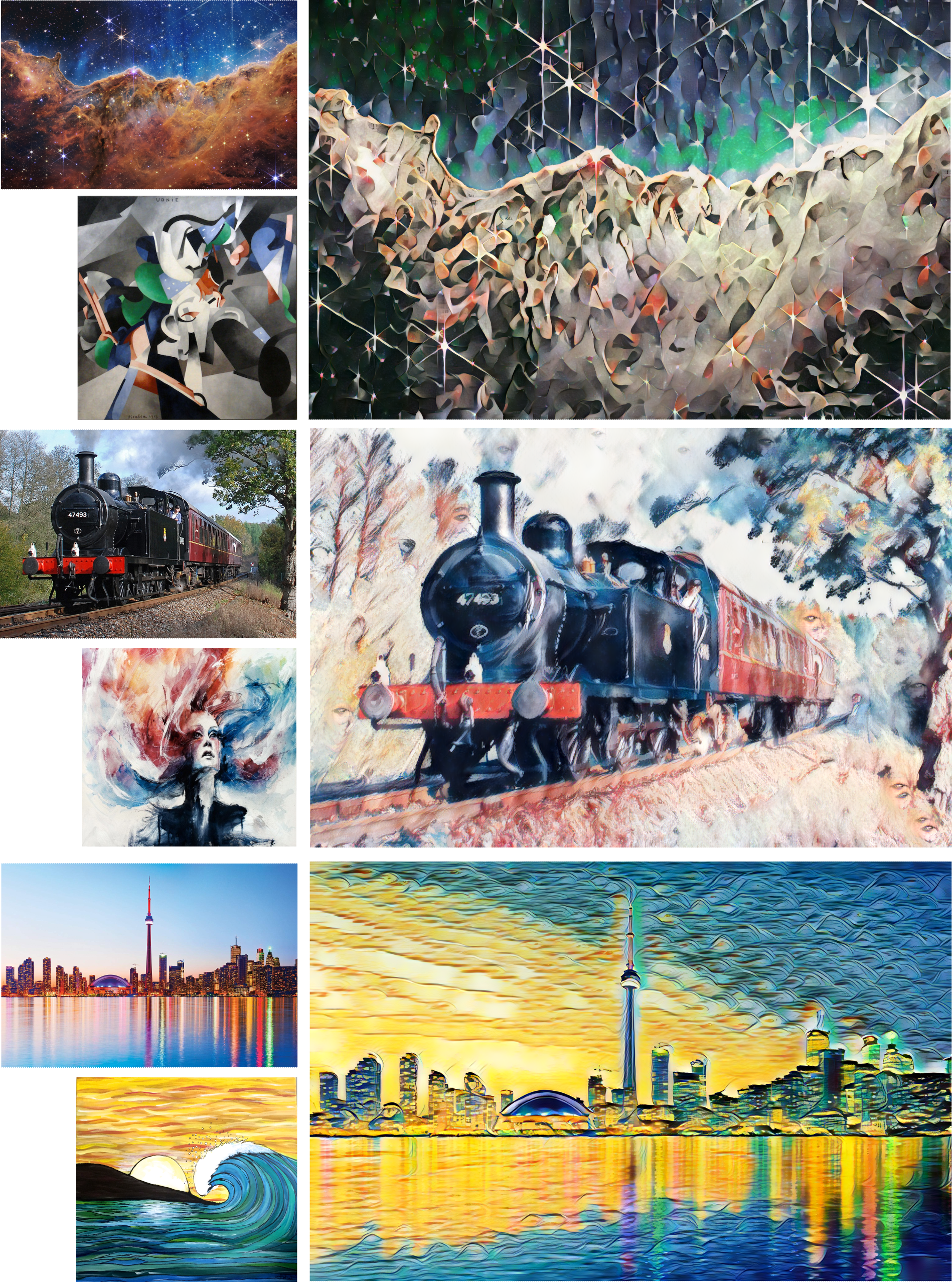} 
    \caption{}
    \label{fig:extra3}
\end{figure*}

\end{document}